\documentclass[10pt,twocolumn,letterpaper]{article}

\usepackage{wacv}
\usepackage{times}
\usepackage{epsfig}
\usepackage{graphicx}
\usepackage{amsmath}
\usepackage{amssymb}
\usepackage{bbm}
\usepackage{algorithmicx}
\usepackage[dvipsnames]{xcolor}
\usepackage{soul}
\usepackage[ruled,vlined]{algorithm2e}
\usepackage{authblk}
\usepackage{float}

\def\wacvPaperID{1184} 

\wacvfinalcopy 

\usepackage[breaklinks=true,bookmarks=false]{hyperref}
\hypersetup{colorlinks}

\begin{document}

\title{HHP-Net: A light Heteroscedastic neural network\\ for Head Pose estimation with uncertainty}

\author[1, 2]{Giorgio Cantarini} 
\author[1]{Federico Figari Tomenotti} 
\author[1]{Nicoletta Noceti} 
\author[1]{Francesca Odone} 

\affil[1]{MaLGa-DIBRIS, Universit\`a degli Studi di Genova, via Dodecaneso 35, 16146-IT Genova,Italy}

\affil[2]{IMAVIS srl, via Trento 5/2, 16145-IT Genova,Italy}

\affil[ ]{\tt\small giorgio.cantarini@imavis.com,
federico.figaritomenotti@edu.unige.it, nicoletta.noceti, francesca.odone@unige.it}

\maketitle

\ifwacvfinal
\thispagestyle{empty}
\fi

\begin{abstract}
In this paper we introduce a novel method to estimate the head pose of people in single images starting from a small set of head keypoints. To this purpose, we propose a regression model that exploits keypoints computed automatically by 2D pose estimation algorithms and outputs the head pose represented by yaw, pitch, and roll. Our model is simple to implement and more efficient with respect to the state of the art -- faster in inference and smaller in terms of memory occupancy --  with comparable accuracy.\\
Our method also provides a measure of the heteroscedastic uncertainties associated with the three angles, through an appropriately designed loss function;  we show there is a correlation between error and uncertainty values, thus this extra source of information may be used in subsequent computational steps. As an example application, we address social interaction analysis in images: we propose an algorithm for a quantitative estimation of the level of interaction between people, starting from their head poses and reasoning on their mutual positions. 
The code is available at \href{https://github.com/cantarinigiorgio/HHP-Net}{\color{magenta}{https://github.com/cantarinigiorgio/HHP-Net}}.

\end{abstract}

\begin{figure*}
\centering
  \includegraphics[width=.85\textwidth]{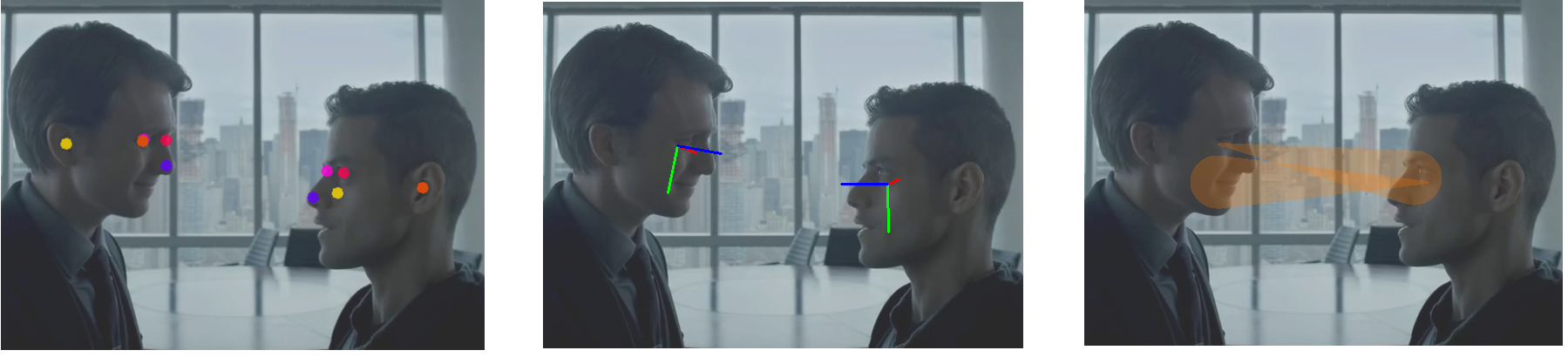}
  \caption{Pipeline of the proposed method: starting from the keypoints detected on the head by a human pose estimation method (left), our method provides the 3D head pose as a triplet of angles yaw, pitch and roll (center) and a measure of uncertainty associated with each angle in the triplet (right, for clarity we only derived a projection on the image plane from yaw and pitch using the Tait-Bryan angles, and represent the uncertainty as a cone around the estimated direction).}
  \label{fig:teaser}
\end{figure*}

\section{Introduction}

Nowadays, 2D human pose estimators are growing in popularity, likewise the number of effective algorithms available in the literature \cite{cao2017realtime, DBLP:conf/iccv/MartinezRIXJSS19, Duan_2019_ICCV, papandreou2018personlab}, since they provide a richer output than simple people detection. Their potential applications are countless, ranging from motion analysis and action recognition, to human-machine interaction, and social interaction analysis (for instance \cite{Shi_2019_CVPR,9007695,9412346}).

A key element in several of the aforementioned applications is the estimation of the head direction \cite{Madrigal2020,abele1986functions,dias2020gaze,GROSSI202061}, that 
aims at computing the pose of human heads with respect to a reference (frontal) pose. 
When observing human agents with conventional cameras, eyes may be difficult to detect, and head direction is often used as a proxy to gaze, knowing that the eyeball orientation can differ by $\pm 35^{^{\circ}}$ degree from the head orientation \cite{Stahl}. \\
\noindent In this paper we estimate the head direction only relying on the sparse and meaningful semantic features provided by pose estimators, with no need of additional input (Figure \ref{fig:teaser}).

We formulate the problem as a multi-task regression and design a Heteroscedastic Neural Network to estimate the head pose expressed in Euler angles.
Thanks to the use of an appropriately designed loss function, the method also associates an uncertainty value -- learned from the data -- with each angle. The concept of uncertainty provides an additional cue that may help the interpretation of the output of the network. 
We estimate aleatoric heteroscedastic uncertainty, that is uncertainty due to data and varying on different inputs. 
 Our head pose estimation method, with a negligible additional effort in terms of space and time resources, can be seen as a {\em plug-in to any given pose estimator}, and allows us to extract precise (in line with more complex state of the art algorithms) head poses.
Indeed, as a positive side effect of operating on a very compact input, the architecture we propose is small in size (it occupies less than 0.5 MB), with a potential to run on mobile architectures, and performs in real time (at about 100 fps).

As an example application of our estimates, we consider social interaction analysis: in particular we propose a light method based on the estimated head poses, for detecting  pair of people looking at each other in images \cite{marin2019laeo, marin2020laeo}. 

In summary, the main contributions of the paper are thus the following:
\begin{itemize}
    \item We introduce a very light (in space and time) head pose estimation method to be used as a plug-in for 2D human pose estimators in RGB images and able to associate an uncertainty with each estimated (pose) angle; the estimated uncertainties represent a cue for model interpretation and strongly correlate with the estimation error.
    \item As a core element of the method, we propose a multi-task regression loss where the uncertainties act as weights of each sub-loss responsible for the estimation of each angle. To the best of our knowledge this is the first attempt to  adopt the concept of uncertainty in multi-task regression for head pose estimation.  
    \item  We provide a thorough  experimental assessment showing how our method is  ``ready" to use for mobile devices, has a space occupancy $\sim$12 times smaller than state-of-the-art while it provides comparable results, in the worst case with about 2 degrees of degradation.
    \item We present a proof-of-concept of the estimated head poses in detecting people looking at each other, as a first step for social interaction analysis; in this context the uncertainty contributes to obtain effective performances.
\end{itemize}

The remainder of the paper is organized as follows: Section \ref{related_work} covers related works on head pose estimation in images and uncertainty evaluation; Section \ref{HHP-Net} presents the proposed HHP-Net architecture, Section \ref{experiments} focuses on the method assessment and the comparison with state-of-the-art. Section \ref{an_application_to_social_interaction_analysis} is about the application to mutual interaction, while Section \ref{discussion} is left to a final discussion.

\section{Related Work}
\label{related_work}

\noindent \paragraph{Head pose estimation methods.} Head pose estimation has been addressed by a number of relatively recent methodologies \cite{10.1109/TPAMI.2008.106}, with classical applications to Human-Machine Interaction or to social interaction analysis. \\
Some methods use additional information such as depth \cite{10.1007/978-3-642-23123-0_11,4405d0521f6748d5a00f4bf4bbe38d40} or time \cite{8099650}, but in this paper we will only refer to methods applicable to RGB images.

Pose can be derived by an estimation of a 3D model \cite{FDGF12},  here we  mention recent deep learning based methods, such as 3DDFA \cite{Zhu_2019}, a CNN able to fit a 3D model to an RGB image; FAN \cite{Bulat_2017} is a state of the art facial landmark detection method, also estimating pose. These approaches propose complex computational pipelines and may obtain rather accurate results.
One of the most recent challenges is in estimating  pose directly from individual  2D images.
On this respect, we start by mentioning a different but related task of estimating the 2D gaze.  GazeFollow \cite{recasens2015were} is a two-pathway CNN architecture that estimates the apparent direction of gaze and the object being observed; it combines saliency maps of the whole image with the position of subjects' head to obtain a pose prediction. A very efficient strategy to provide an estimate of the apparent direction of gaze is proposed in \cite{dias2020gaze}: similarly to our work, the input is obtained by a 2D pose estimator.\\
3D head pose from images has been addressed in several works \cite{drouard:hal-01163663,Lathuiliere_2017_CVPR}, and nowadays it is often obtained by deep learning architectures that start from the output of face detectors: Shao et al \cite{8a49fe5a5a75448fbae7880f4bf96c72} propose an adjustment of the ROI obtained by face detection (it incorporates an offset around the face) and  a combined regression and classification loss. HopeNet is a regression method with ResNet and a joint MSE and cross-entropy loss \cite{ruiz2018finegrained}.
FSA-net \cite{Yang_2019_CVPR} is a two-stream multi-dimensional regression network able to provide accurate fine grained estimations.\\
We propose a 3D head pose estimation from RGB images, but instead of estimating pose directly from the image, we rely on the output of a human pose detector, similarly to the strategy proposed in \cite{dias2020gaze} for 2D heading estimation. This allows us to design a very light architecture, still able to achieve accurate results.\\
In this section it is also worth mentioning multi-task approaches, exploiting the concept of training an architecture to achieve joint results, an approach proved to improve performances: KEPLER \cite{7961750} predicts facial keypoints and pose with a modified GoogLeNet; a coarse pose is used to improve keypoints detection.
Hyperface \cite{8170321} simultaneously performs face and landmark detection, pose estimation and gender recognition. A very recent paper \cite{cao2020vectorbased} proposes a reformulation of the problem in terms of a rotation matrix to be used to formulate the output. 
\paragraph{ Uncertainty estimation.} As recently observed in \cite{zhou2020whenet}, where the WHENET architecture is proposed, head pose estimation is intrinsically harder on certain view-points. This paper extends \cite{ruiz2018finegrained}, by changing the loss functions specifically addressing wide-range head pose, to perform well on lateral views.
 Instead our work follows the observation in \cite{dias2020gaze}: certain view-points are associated with  different levels of uncertainty, creating a large discrepancy in accuracy.  This can be formalized with the concept of aleatoric heteroscedastic uncertainty \cite{NIPS2017_2650d608}, which depends on the inputs, and may be estimated from data. Conventional deep learning methods are unable to estimate the uncertainty of their inputs, for this Bayesian deep learning is becoming very popular on this respect \cite{8578879,BMVC2017_57,prokudin2018deep}. In our method we propose a multi-task approach where a task is associated with one of the three pose angles, extending \cite{NIPS2017_2650d608} to the multi-loss case.

 Indeed \cite{8578879} reports a loss with homoscedastic uncertainty, also called task-dependent uncertainty, that is constant across different inputs. In this way, their model can learn the weight of each task.

\section{The proposed method: HHP-Net}
\label{HHP-Net}

The starting point of our approach is the output of a pose detector (the literature today offers various alternatives, as OpenPose \cite{cao2017realtime, DBLP:conf/iccv/MartinezRIXJSS19}, CenterNet \cite{Duan_2019_ICCV}, and PoseNet \cite{papandreou2018personlab}) providing a set of keypoints roughly describing the pose of a human body in an image. These detectors commonly provide also a confidence on the keypoint location estimate, which represents an additional source of knowledge that can be injected in our approach.

We model the estimation of the head orientation as a multi-task regression problem, where a Neural Network predicts the 3D vector of the head orientation with angles in Euler notation  ({\it yaw, pitch} and {\it roll}).
The input is formed by a set of $n$ semantic keypoints located on the image plane: $ \{(x_1^i, x_2^i, c^i)\}_{i=1}^n$, with $x_1^i$ the horizontal and  $x_2^i$ vertical coordinates and  $c_i$ the confidence of the $i$-th keypoint.

Coordinates are centered with respect to their centroid and then normalized with respect to their corresponding maximum value; $c_i$ is provided in the range $[0,1]$. 
The value of confidence is particularly important, as it encodes missing points ($c=0$) and low confidence points. These situations may occur frequently in real-world applications, in particular in human-human interaction, because of occlusions, self occlusions or lateral poses.   
\begin{figure}[htb]
  \centering
  \includegraphics[width=1\linewidth]{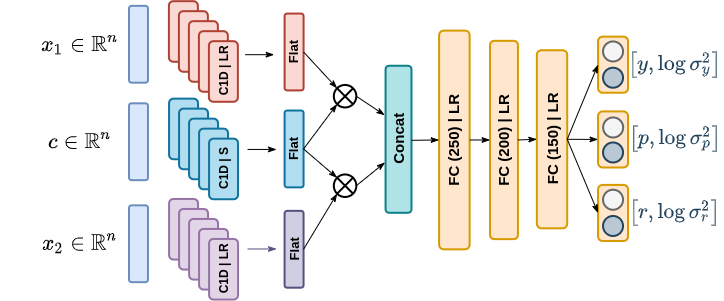}
  \caption{The light architecture of our approach.  C1D = 1D Convolution, S = sigmoid, LR = leakyReLu, $\bigotimes$ = element-wise multiplication
}
  \label{fig:architecture}
\end{figure}

\subsection{The architecture}

Figure \ref{fig:architecture} provides a sketch of our architecture. We formalize the input of the network as a triplet of vectors $\mathbf{x}_1 = [x_1^1, \dots, x_1^n]$, $\mathbf{x}_2 = [x_2^1, \dots, x_2^n]$ and $\mathbf{c} = [c^1, \dots, c^n]$. The input vectors are first processed in independent streams, with a 1D Convolutional layer followed by a non-linear activation:  Leaky ReLU, to avoid vanishing gradient issues,  for $\mathbf{x}_1$ and $\mathbf{x}_2$, and sigmoid activation for the confidence vector $\mathbf{c}$, to smoothly control  the impact of different confidence values. After the convolution and the non-linear activation, from $x_1$ $x_2$ and $c$ we obtain respectively $x^{\ast}_1$, $x^{\ast}_2$ and $c^{\ast}$.
 
The latter are flattened and combined, using the element-wise multiplication to obtain two vectors $v_1 = x^{\ast}_1 \otimes c^{\ast}$ and $v_2 = x^{\ast}_2 \otimes c^{\ast}$, following the logic of the Confidence Gated Unit (CGU) proposed in \cite{dias2020gaze}.

The two gated outputs $v_1$ and $v_2$ are concatenated to obtain a single vector, which is provided to the intermediate part of the architecture, where a sequence of three fully connected layers consisting of  250, 200 and 150 neurons respectively is employed. Each layer includes a LeakyReLU, again to avoid vanishing gradients, as a non-linear activation function.
Three output layers return the estimated angles, each of which is associated with its uncertainty value.

\begin{figure*}[ht]
    \centering
    \includegraphics[width=.18\linewidth, height=.20\linewidth]{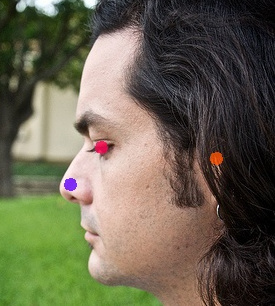}
    \includegraphics[width=.18\linewidth, height=.20\linewidth]{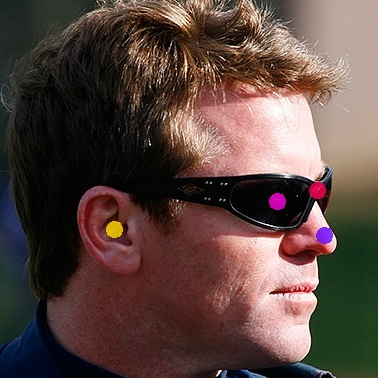}
    \includegraphics[width=.18\linewidth, height=.20\linewidth]{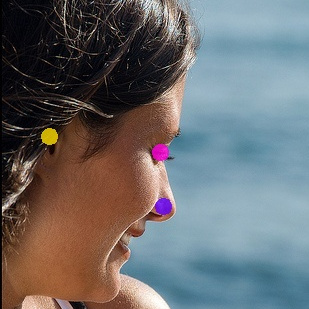}
    \includegraphics[width=.18\linewidth, height=.20\linewidth]{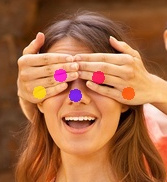}
    \includegraphics[width=.18\linewidth, height=.20\linewidth]{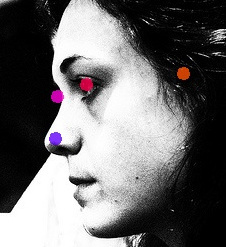}
    \\ 
    \includegraphics[width=.18\linewidth, height=.20\linewidth]{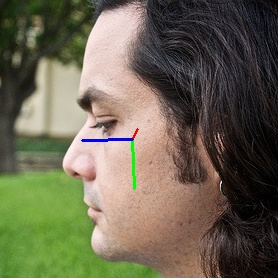}
    \includegraphics[width=.18\linewidth, height=.20\linewidth]{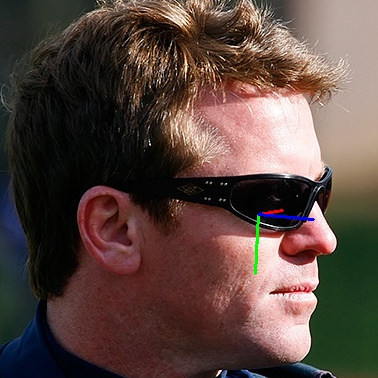}
    \includegraphics[width=.18\linewidth, height=.20\linewidth]{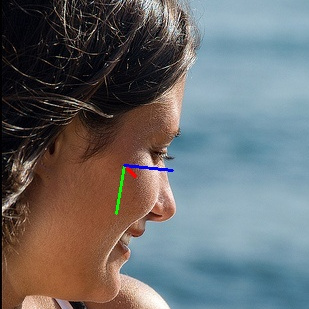}
    \includegraphics[width=.18\linewidth, height=.20\linewidth]{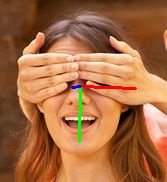}
     \includegraphics[width=.18\linewidth, height=.20\linewidth]{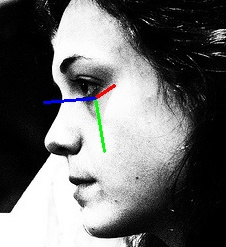}
    \caption{Examples on the AFLW2000 Dataset. Top: keypoints input of our estimate; bottom: estimated head pose.}
    \label{fig:samples}
\end{figure*}

\subsection{Estimating the 3D head orientation}
For training the network we adopt a loss function incorporating heteroscedastic aleatoric uncertainty. With respect to classical Neural Networks, an Heteroscedastic Neural Network provides an estimate of the uncertainty of each prediction. This is particularly useful to capture noise within input observations: noise in our case is related with inherent keypoints detection which may be affected by difficult viewpoints or occlusions. Indeed, some poses are intrinsically noisier and more prone to self-occlusions.

This type of uncertainty may be learned as a function of the data, thus the output will include not only  the three angles yaw, pitch, roll, stored in a vector $\mathbf{q}=[y,p,r]$, but also the uncertainty values associated with them $\mathbf{s}=[s_y,s_p,s_r]$.

We use a multi-task loss function $\mathcal{L}_{HHP}$\footnote{The derivation of the loss is in the Supplementary Material \ref{derivation_loss}.} for training our network, defined as follows:

\begin{equation*}
   \sum_{i \in \{y,p,r\}} \left ( \frac{1}{2} \exp \left( -s_{i} \right) \left\| q_{i} - f_{i}\left( \mathbf{x}_1, \mathbf{x}_2,\mathbf{c}  \right) \right\|^{2} + \frac{1}{2} s_i  \right)
\end{equation*}

where $f_i$ is the $i-th$ component (associated with either yaw, pitch, or roll) of $\mathbf{f}$, the estimate obtained by the Heteroscedastic Neural Network, and 
$s_{i} = \log \sigma_i(\mathbf{x}_1, \mathbf{x}_2,\mathbf{c} )^{2} $ 
where $\sigma_i^{2}$ 
is the variance of a normal distribution we assume to have generated the noise \cite{nix1994}:
learning the logarithm of the variance allows us to obtain a more stable solution, avoiding potential divisions by zero  \cite{NIPS2017_2650d608}.

With this formulation we obtain a data-driven uncertainties estimation for each angle, used as a weight of each sub-loss. The uncertainty can increase the robustness of the network when dealing with noisy input data, in fact we will empirically show a correlation between uncertainty and estimation error

\section{Experiments}
\label{experiments}

\subsection{Implementation details}

In this work we adopt OpenPose \cite{cao2017realtime} as a keypoints extractor, as it provides a good balance between efficiency (it performs in real-time) and effectiveness. Among the 25 body keypoints provided by OpenPose, in this work we focus on the five located on the face --  left and right eye, left and right ear, nose -- thus obtaining a triplet of input vectors $\mathbf{x}_1 = [x_1^1, \dots, x_1^5]$, $\mathbf{x}_2 = [x_2^1, \dots, x_2^5]$ and $\mathbf{c} = [c^1, \dots, c^5]$.\\
For the initialization, the weights of each layer are randomly sampled from a normal distribution with $\mu=0$ and $\sigma^2 = 0.05$.
The network has been trained for a number epochs that ranges from 100 to 1000 depending on the dataset using Adam as optimizer, with learning rate $0.001$, and batch size of 64. The weights associated with the best validation loss have been selected as final model.

\subsection{Datasets and protocols}
We evaluate the effectiveness of our approach on three different datasets: 
\begin{itemize}
\item{BIWI} \cite{fanelli2011real} includes $\sim15K$ images of 24 people acquired in a controlled scenario. The head pose orientation covers the range $\pm 75^o$ for the yaw angle, and $\pm 60^o$ for the pitch. The ground truth has been obtained by fitting a 3D face model.
\item AFLW-2000 \cite{yin2015} contains the first 2000 images of the in-the-wild AFLW dataset \cite{koestinger11a}, a large-scale collection of face images with a large variety in appearance and environmental conditions. The annotation has been obtained by fitting a 3D face model.
\item{300W-LP} \cite{sagonas2013300} is a collection of different in-the-wild datasets, grouped and re-annotated to account for different types of variability, as  pose, expression, illumination, background, occlusion, and image quality. A face model is fit on each image, distorted to vary the yaw of the face.
\end{itemize}

According to previous works \cite{ruiz2018finegrained,Yang_2019_CVPR,8a49fe5a5a75448fbae7880f4bf96c72}, in the comparative analysis we adopt two main protocols:
\begin{itemize}
    \item[P1] Training is performed on a single dataset (300W-LP), while BIWI and AFLW-2000 are used as test.
    \item[P2] Training and test set are derived from the BIWI dataset using the split 16-9 sequences, for training and test respectively, following the procedure proposed in \cite{FDGF12}.
\end{itemize}

\subsection{Method assessment}
\label{method_assessment}

We start the discussion on the experimental analysis showing samples of qualitative results in Figure \ref{fig:samples}, where we derived the directions on the image plane according to the Tait-Bryan angles. In spite of the sparseness of the input representation (keypoints overlaid on the original image, top row), the estimated pose is very accurate on a variety of challenging conditions (bottom row). 

In the following we provide an assessment to discuss properties and meaningfulness of the uncertainty measures.\\

\begin{figure}[htb]
  \centering
   \includegraphics[width=.48\linewidth]{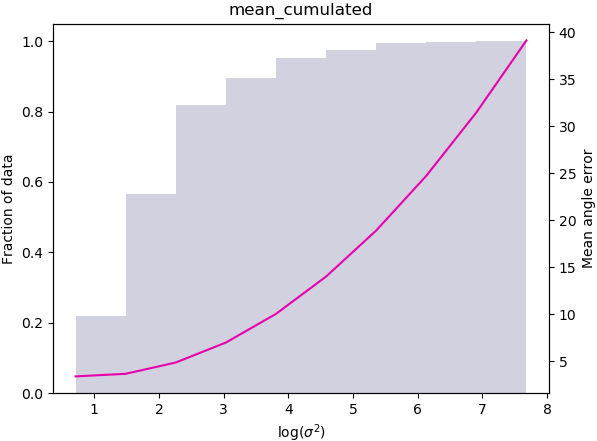}
    \includegraphics[width=.48\linewidth]{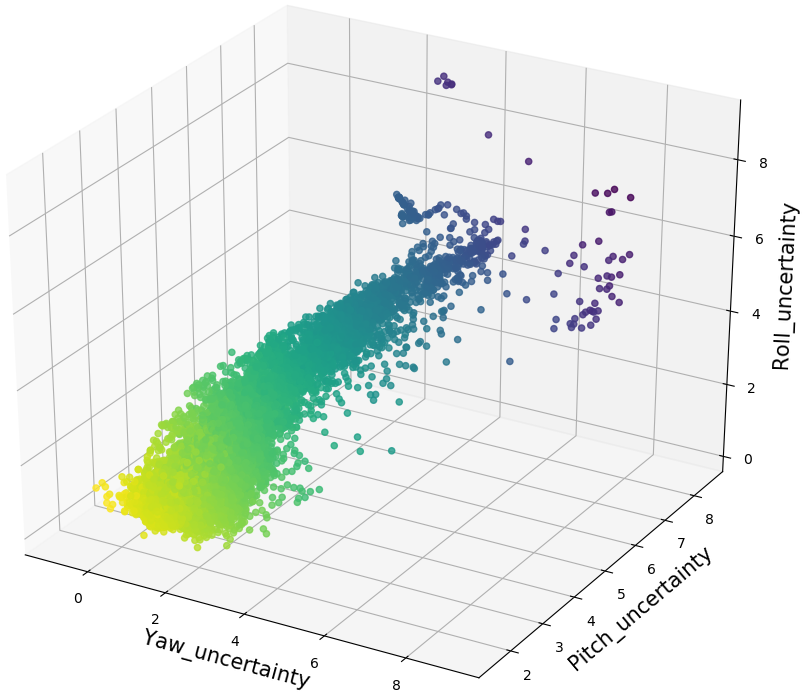}
  \caption{ Left: Cumulative  angular  error  as  a  function  of  the average uncertainty. Right: Linear correlation among the three uncertainties  (Pearson correlation between (uncertainties of) {\it yaw} and {\it pitch} is 0.72, the one between  {\it yaw} and {\it roll} is 0.78, the one between {\it pitch} and {\it roll} is 0.92)}
  \label{fig:cumulative_mean_err_unc}
\end{figure}

\noindent {\bf On the quality of the uncertainty estimations}.
We report some qualitative experiments to highlight the properties of the uncertainty measures incorporated in our model.
Figure \ref{fig:cumulative_mean_err_unc} (left) reports a cumulative analysis on the amount of data associated with a given uncertainty, highlighting how the average error grows with the uncertainty -- in agreement with \cite{dias2020gaze}.
Similarly to what observed in \cite{feng2019leveraging}, we also notice a strong correlation between the uncertainty values associated with the three predicted angles: in Figure \ref{fig:cumulative_mean_err_unc} (right) we provide a visual representation where a linear correlation between the uncertainties associated with the three predictions can be easily appreciated.\\ 

\begin{figure}[htb]
  \centering
  \includegraphics[width=1.0\linewidth]{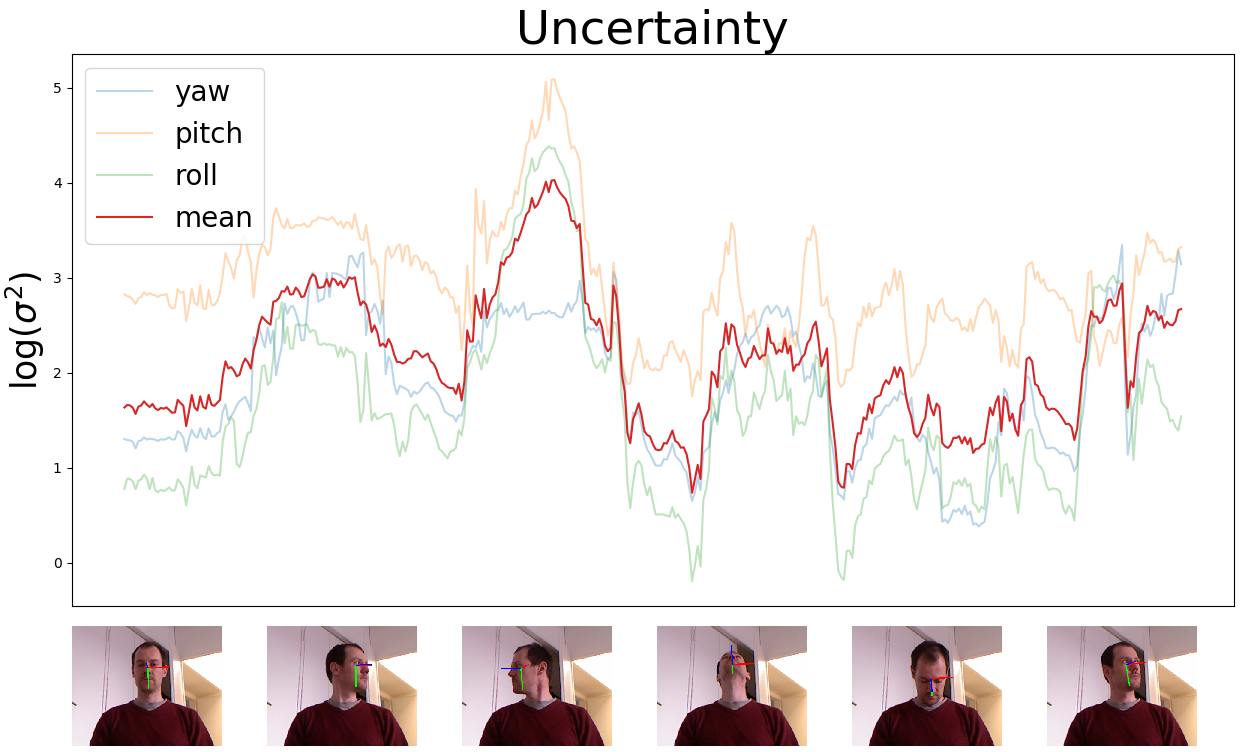}
  \caption{Uncertainty values are influenced by the head pose; the red line is the mean between the three uncertainties associated to each angle.}
  \label{fig:qualitative_video}
\end{figure}

\noindent {\bf Uncertainty estimation and model interpretation.} 
In Figure \ref{fig:qualitative_video} we report the trend of the uncertainty associated with the prediction obtained from a video sequence where a subject rotates the head offering different test poses to the method.
Representative frames --  that provide an intuition about the transitions between poses in the sequence -- are reported below the plot. It is easy to observe that for some poses -- the ones associated with ambiguous views or partial occlusions that hide some keypoints on the face -- the uncertainty tends to be higher. The lowest uncertainty values are associated with frontal views, the ones providing the most visible and non-ambiguous keypoints.\\
Inspired by these observations, we now evaluate the dependence of the uncertainty and the error on the quality and quantity of the input keypoints.\\

\begin{figure}[H]
    \centering
    \includegraphics[width=.48\linewidth]{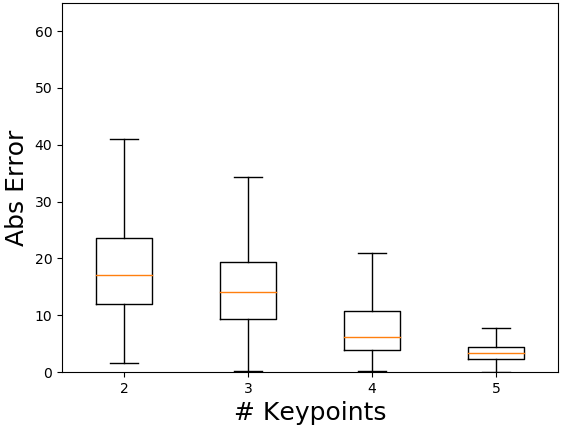}
    \includegraphics[width=.48\linewidth]{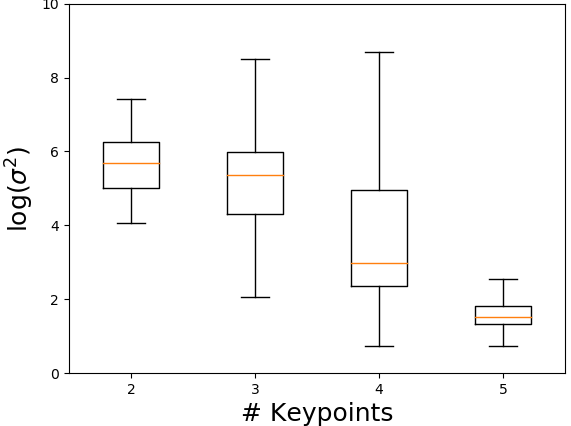}
    
    \caption{Performance of our method (left: mean angular error, right: uncertainty) with respect to the number of input points provided in input. Training: 300W-LP, Test: BIWI (plots refer to the latter).} 
    \label{fig:boxplot}
\end{figure}

\noindent{\bf On the number of keypoints} 
In Figure \ref{fig:boxplot} we analyse the performance of our method in terms of uncertainty values (right) and absolute angular error (left) as we group the input data according to the number of keypoints provided in input. 

Since in the worst case OpenPose provides at least  three keypoints for the dataset used in this experiment, we randomly dropped points from the input to evaluate the behavior of the method in more challenging scenarios. 

When all  the 5 keypoints are available, the uncertainty is compactly lower 
as the method can rely on a more comprehensive representation of the input. In the intermediate cases -- where we may have 2, 3, or 4 keypoints available in input -- the uncertainty progressively decreases, but we also have a higher degree of variability, as some keypoints configurations are more significant than others and thus the amount of information they provide to the model may be uneven, also reflecting the concept that the noise could be different for each input sample. \\

\noindent{\bf Removing the uncertainty: an ablation study}
In this final experiment, we perform an ablation study by removing the uncertainty from our model. To this purpose, we consider  two variations: in the first (MSE) we  directly regress the three angles adopting a loss computed as the sum of the Mean Squared Error  on each angle. In the second (COMB) we  employ an alternative loss function that jointly solves a regression and a classification task, which has been successfully applied to the same estimation task \cite{ruiz2018finegrained}. \\
In Table \ref{tab:lossfunction} we report the angular errors we obtain with the three different loss functions. As it can be observed, learning the angles associated with the uncertainty provides the best average performance\footnote{In the Supplementary Material \ref{remove_unv} details on the experiments are reported.}.\\ 

\begin{table}[h]
    \centering
    \caption{Comparison among different loss functions (see text). {\bf All errors are expressed in degrees ($^\circ$)}:  
   MAE = Mean Absolute Error (the subscript refers to the loss)}
    \smaller{
    \begin{tabular}{|c|c||c|c|c|}
    \hline
\textbf{Train} & \textbf{Val} & \textbf{MAE$_{MSE}$} &  \textbf{MAE$_{COMB}$} & \textbf{MAE$_{UNC}$} \\
    \hline\hline
    BIWI & BIWI & 3.70 & 3.80 & {\bf3.68}\\ 
     \hline
     300WLP & BIWI & 5.28 & 5.88 &  {\bf 5.18} \\ 
     \hline
     300WLP & AFLW2000 & 8.07 & 8.04 & {\bf 7.70} \\ 
     
     \hline
     AFLW & AFLW2000 & 6.26 & 6.18 & {\bf  6.16} \\ 
     \hline

    \end{tabular}}
    \label{tab:lossfunction}
\end{table}

\subsection{Comparisons with other approaches}
We now perform a comparative analysis with state of the art head pose estimators. For a fair comparison, we consider methods that use RGB images as inputs {or features extracted from them}. We always report the performance provided by our method in its largest variant (0.4MB) and we apply the same rule for alternative approches.\\
The analysis is reported in Table \ref{tab:1}, \ref{tab:2}, and \ref{tab:3}. As a first important observation, notice that our approach produces a significantly smaller model (0.4 MB).  This was the main purpose of our work and it has been clearly achieved, as our method is about $\sim$12 times smaller than the closest model in literature.
According to the protocol followed by other works -- all requiring a face detector but  not including its size in their analysis\footnote{as benchmark datasets provide the face bounding boxes.} -- the size of our model does not include the pose estimator. 

In terms of performances, Table \ref{tab:1} reports a comparison with respect to Protocol P2 (BIWI dataset for training and test): the results we obtain are superior to \cite{4405d0521f6748d5a00f4bf4bbe38d40,drouard:hal-01163663, FDGF12} and slightly below  \cite{8099650,Lathuiliere_2017_CVPR,Yang_2019_CVPR, DBLP:conf/aaai/ZhangWLY20} (less than 0.1 degree of difference for the first three, less than 0.4 for the latter).

Table \ref{tab:2} refers to Protocol P1 (training carried out on 300W-LP, BIWI for test): the experiment mainly evaluates the transfer potential to a different dataset with different properties. 
The table reports results obtained with methods relying on the estimation of 3D face models \cite{Zhu_2019,7961750,kazemi2014one,Bulat_2017} and methods based on analysing RGB image portions obtained by face detectors, such as \cite{8a49fe5a5a75448fbae7880f4bf96c72,ruiz2018finegrained,Yang_2019_CVPR}. 

\begin{figure}
    \centering
    \includegraphics[width=.31\linewidth]{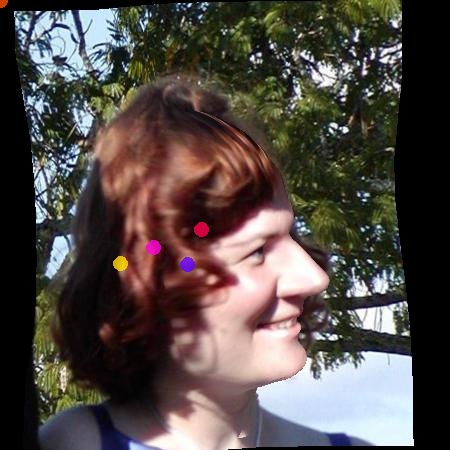}
    \includegraphics[width=.31\linewidth]{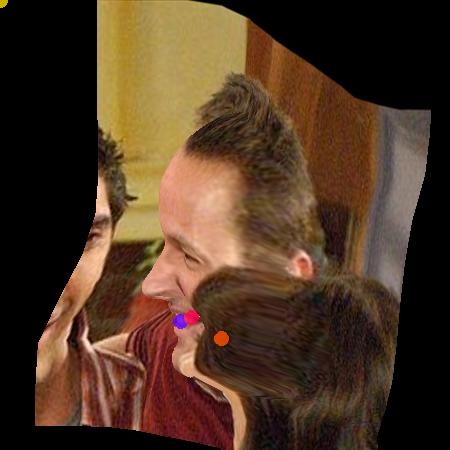}
    \includegraphics[width=.31\linewidth]{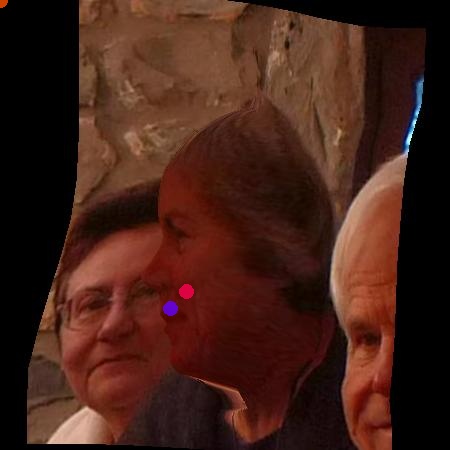}
    
    \caption{Examples of failures in the keypoints detection on the 300W-LP dataset caused by the artifacts due to the synthetic data manipulation; the points are clearly in the wrong positions.}
    \label{fig:synth}
\end{figure}

We share with the latter group the main motivation of designing simple and more efficient procedures while keeping competitive performances. In this sense, our approach does not require complex pre-processing steps or highly resource-demanding training, but at the same time it wisely leverages structural information on the face. Table \ref{tab:2} reports results that are more accurate than all methods with the exception of FSA-Caps, although the difference is on average only slighlty above 1 degree. This small accuracy loss is counterbalanced by  the benefits in terms of a smaller size, and it may be explained by the simplicity and compactness of our input: while nicely behaving in the majority of non-ambiguous situations, our sparse input is more severely influenced by occlusions, and missed or noisy detections. \\
Finally, Table \ref{tab:3} follows again Protocol P1, on a more complex test set (as images are acquired in a less controlled environment). In this case our methodology is reporting slightly worse results,  but with a loss always less that 3 degrees on average. We noticed this is due in particular to keypoint detection errors, as the the synthetic data manipulation (see examples in Figure \ref{fig:synth}) introduced artifacts.
To further evaluate the transfer potential of our approach we also report the result we obtained on the same test set, when training the network on a related dataset (AFLW without the AFLW2000 section): the results are in this case comparable to the previous experiments.\\

We conclude by mentioning that, among the very recent approaches, our comparative analysis does not include WHENET \cite{zhou2020whenet} since it addresses specifically wide-angle cases and thus it adopts a different training set, not available at the moment, in its experimental analysis;  its space occupancy is not reported in  the paper, although the implementation they provide is 18 MB.
We also do not report a comparison with \cite{cao2020vectorbased}, since their problem definition and their evaluation is not immediately comparable with state of the art.

\begin{table}
  \caption{Comparison following Protocol P2: BIWI train-test.
  {\bf All errors are expressed in degrees ($^\circ$)}:  
    err$_y$= yaw error, err$_p$=pitch error, err$_r$= roll error. MB = Models size in megabytes.}
  \label{tab:1}
  \small{ 
  \begin{tabular}{|c || c || c | c | c | c |}
    \hline
    \textbf{Method} & \textbf{MB} & \textbf{err$_y$} & \textbf{err$_p$} & \textbf{err$_r$} & \textbf{MAE}\\
    \hline\hline
    D-HeadPose \cite{4405d0521f6748d5a00f4bf4bbe38d40} & -  & - & 5.67 & 5.28 & - \\ 
     \hline
     Drounard et al \cite{drouard:hal-01163663}& -  & 4.9 & 5.9 & 4.7 & 5.16 \\
     \hline
    Fanelli et al.\cite{FDGF12} & - & 3.8 & 3.5 & 5.4 & 4.23 \\
     \hline
    DFA\cite{8099650} & 500  & 3.91 & 4.03 & 3.03 & 3.66 \\ 
     \hline
    DMLIR \cite{Lathuiliere_2017_CVPR} & 500  & 3.12 & 4.68 & 3.07 & 3.62 \\ 
     \hline
    FSA-Net \cite{Yang_2019_CVPR} & 5.1 &  2.89 & 4.29 & 3.60 & 3.60 \\ 
    \hline
     FND \cite{DBLP:conf/aaai/ZhangWLY20} & 5.8 & 3.0 & 3.98 & 2.88 & 3.29 \\
     \hline
     \hline

     \textbf{Our approach} & {\bf 0.4}  & 3.04 & 4.79 & 3.21 & 3.68 \\ 
     
  \hline
\end{tabular}
}
\end{table}

\begin{table}
   \caption{Comparison following Protocol P1: 300W-LP train, BIWI test.
  {\bf All errors are expressed in degrees ($^\circ$)}:  
    err$_y$= yaw error, err$_p$=pitch error, err$_r$= roll error, MAE = Mean Absolute Error. { MB = Models size in megabytes.}}
  \label{tab:2}
  \small{
  \begin{tabular}{|c || c || c | c | c | c |}
    \hline
    \textbf{Method} & \textbf{MB} & \textbf{err$_y$} & \textbf{err$_p$} & \textbf{err$_r$} & \textbf{MAE}\\
 \hline\hline
    3DDFA \cite{Zhu_2019}& -  & 36.2 & 12.3 & 8.78 & 19.1 \\ 
     \hline
     KEPLER \cite{7961750} & -  & 8.8 & 17.3 & 16.2 & 13.9 \\
     \hline
    Dlib \cite{kazemi2014one} & -  & 16.8 & 13.8 & 6.19 & 12.2 \\
     \hline
     FAN \cite{Bulat_2017} & 183  & 8.53 & 7.48 & 7.63 & 7.89 \\
     \hline
     Shao(K=0.5)\cite{8a49fe5a5a75448fbae7880f4bf96c72} & 93   & 4.59 & 7.25 & 6.15 & 6.00 \\
     \hline
    Ruiz \cite{ruiz2018finegrained}($\alpha$=2) & 95.9  & 5.17 & 6.98 & 3.39 & 5.18 \\
     \hline
      FSA \cite{Yang_2019_CVPR} & 5.1   & 4.27 & 4.96 & 2.76 & 4.00 \\
      \hline\hline

      \textbf{Our approach} & {\bf 0.4}  & 4.14 & 7.00 & 4.40 & 5.18 \\

     \hline
\end{tabular}
}
\end{table}

\begin{table}
  \caption{Comparison following Protocol P1: 300W-LP train, AFLW 2000 test
  ($\dagger$ = Trained on (AFLW - AFLW2000))
   {\bf All errors are expressed in degrees ($^\circ$)}:  
    err$_y$= yaw error, err$_p$=pitch error, err$_r$= roll error, MAE = Mean Absolute Error. { MB = Models size in megabytes.}}
  \label{tab:3}
  \small{
  \begin{tabular}{|c || c || c | c | c | c |}
    \hline
    \textbf{Method} & \textbf{MB} & \textbf{err$_y$} & \textbf{err$_p$} & \textbf{err$_r$} & \textbf{MAE}\\
    \hline\hline
    Dlib \cite{kazemi2014one} & -  & 23.1 & 13.6 & 10.5 & 15.8 \\
     \hline
     FAN \cite{Bulat_2017} & 183  & 8.53 & 7.48 & 7.63 & 7.88 \\
    \hline
    Ruiz\cite{ruiz2018finegrained}($\alpha$=2) & 95.9  & 6.92 & 6.64 & 5.67 & 6.41 \\
     \hline
     Shao(K=0.5)\cite{8a49fe5a5a75448fbae7880f4bf96c72} & 93 & 4.59 & 7.25 & 6.15 & 6.00 \\
      \hline
      FSA \cite{Yang_2019_CVPR} & 5.1  & 4.50 & 6.08 & 4.64 & 5.07 \\
      
      \hline\hline

     \textbf{Our approach } & {\bf 0.4} & 5.26 & 10.12 & 7.73 & 7.70 \\

     \hline\hline
     \textbf{Our approach$^{\dagger}$} & {\bf 0.4}  & 7.40  & 6.63 & 4.47 & 6.16 \\

     \hline
\end{tabular}
}
\end{table}


\section{An application to social interaction analysis}
\label{an_application_to_social_interaction_analysis}

We finally discuss a task where our method finds a natural application, i.e. the analysis of social interaction, for which  the gaze or the head directions represent a strong visual cue of non-verbal human-human communication \cite{abele1986functions}.  We consider scenarios where a small group of people is involved in a social experience, and we pay particular attention to people {\it looking at each other} (LAEO).\\   
 
\begin{figure}[htb]
  \centering
  \includegraphics[width=.8\linewidth]{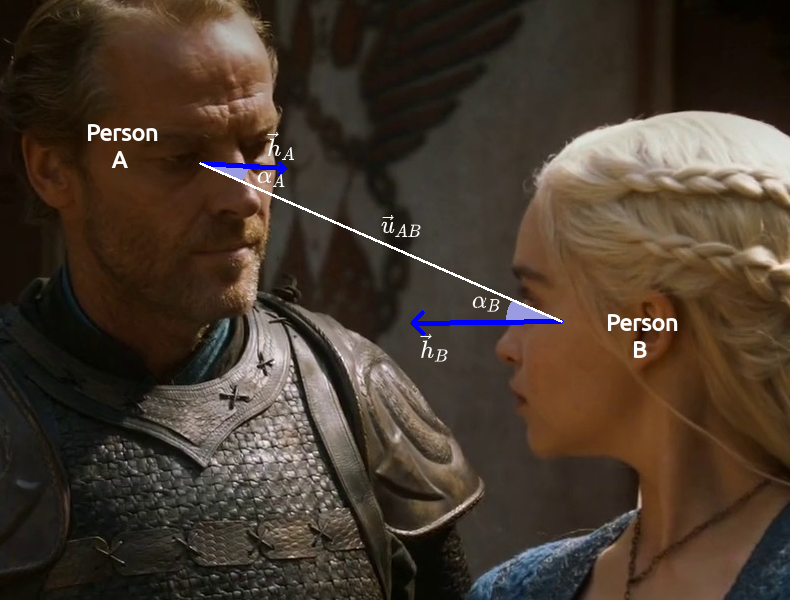}
  \caption{A visual sketch with our formulation of the LAEO detection task (for readability the vectors are denoted with arrows).}
  \label{fig:interaction_analysis}
\end{figure}

\noindent {\bf LAEO algorithm.} Figure \ref{fig:interaction_analysis} provides a visual sketch with our formulation of the task. Let us consider the two people present in the scene, $A$ and $B$ in our example, whose positions can be compactly described with the head centroids $(x_A, y_A)$ and $(x_B, y_B)$. We start from the pose estimated for each of them (3 Euler angles) and obtain a projection of the corresponding direction on the image plane. More in details for the subject $A$, given the triplet of angles $(y_A, p_A, r_A)$, we derive the end-point of the head direction on the image plane $(x_A^{\prime}, y_A^{\prime})$ as  $x_A^{\prime}=\sin(y_A)$ and  $y_A^{\prime}=-\cos(y_A) \sin(p_A)$.\\
Then, we estimate a measure of interaction between each pair of people by proposing a simple but effective method. 
We consider the vector $\mathbf{u}_{AB}$ connecting the two head centroids, the vector $\mathbf{h}_A$ and the angle $\alpha_A$ between the two: the measure of the interaction is given by the cosine of the angle $\alpha_A$. The same applies to person $B$ with $\mathbf{u}_{BA}=-\mathbf{u}_{AB}$ and $\alpha_B$. The average between the two measures gives the LAEO value and a thresholding on such measure allows us to detect LAEO pairs.\\

We build on this baseline method exploiting the knowledge we derive from the uncertainty associated with the 3D angles. Given the triplets of uncertainties associated with the two heads poses, $(s_A^y, s_A^p, s_A^r)$ and  $(s_B^y, s_B^p, s_B^r)$, we compute the averages, $\hat{s}_A = \frac{1}{2}(s_A^y+s_A^p)$ and $\hat{s}_B = \frac{1}{2}(s_B^y+s_B^p)$; the roll component is discarded because it does not affect the gaze vector projection on the image plane.

Following the intuition that estimates with high uncertainty should be less reliable, we compute a weight to adjust the contribution of each subject to the interaction measure depending on the confidence we have in it, essentially deciding a threshold above which the estimate is considered not reliable. For the subject $A$ this can be formulated as $w_A = \mathbbm{1}_X(\hat{s}_A)$ where $X=[0, \delta]$ with $\delta$ an appropriate threshold on the uncertainty, and $\mathbbm{1}_X:\mathbb{R} \rightarrow \{0,1\}$ the indicator function on the interval $X$. $\delta$ is computed as the average uncertainty plus the standard deviation, both of them computed on the entire training set (in the experiments $\delta=7$). The method is sketched in Algorithm \ref{algo_a1}.\\ 

\begin{algorithm}[h]
 \caption{Fast LAEO Detection}
 \begin{algorithmic}[1]
  \State \textbf{Input}: Head centroids $(x_A, y_A)$ and $(x_B, y_B)$; projections of head directions $(x_A^{\prime}, y_A^{\prime})$ and $(x_B^{\prime}, y_B^{\prime})$; uncertainty weights $w_A$ and $w_B$
  \State $\mathbf{u}_{AB} \leftarrow (x_B-x_A, y_B-y_A)$ \;
   \State $\mathbf{h}_{A} \leftarrow (x_A^{\prime}-x_A, y_A^{\prime}-y_A)$ \;
   \State $\mathbf{h}_{B} \leftarrow (x_B^{\prime}-x_B, y_B^{\prime}-y_B)$ \;
   \State $cos(\alpha_A) \leftarrow \frac{\mathbf{u}_{AB} \cdot \mathbf{h}_{A}}{|\mathbf{u}_{AB}|\cdot|\mathbf{h}_{A}|}$ \;
   \State $cos(\alpha_B) \leftarrow \frac{-\mathbf{u}_{AB} \cdot \mathbf{h}_{B}}{|\mathbf{u}_{AB}|\cdot|\mathbf{h}_{B}|}$\;
   \State Compute the level of mutual interaction $ LAEO_{value} = w_A cos(\alpha_A) + w_B cos(\alpha_B)$\;
   
   \State Return $LAEO_{value}$
  
 \end{algorithmic}
 \label{algo_a1}
\end{algorithm}

\noindent {\bf LAEO assessment.} We evaluate our method on the UCO-LAEO dataset \cite{marin2019laeo}, that includes sequences from  
four popular TV shows in the form or 129 shots of variable length. The annotation is provided at a frame level -- {\it is there a pair of LAEO people in the frame?} --  and at a pair level -- i.e. each head pair is labeled as LAEO or not. The task we solve is a binary classification task: for each frame in the sequence we consider all pairs of people detected in the frame and label them as LAEO or not using the method in Algorithm \ref{algo_a1}. Finally a threshold $\tau$, selected on the ROC curve of the training set, is finally used to detect the LAEO pairs.

\begin{table}[h]
    \centering
    \caption{The performance of our method for LAEO detection on the UCO-LAEO dataset. AP is estimated as in \cite{marin2020laeo}, $\tau=0.93$.}
    \begin{tabular}{|l||c|c|c|c|}
    \hline
    {\bf Method} & {\bf PREC} & {\bf REC} & {\bf F} & {\bf AP}\\
    \hline\hline
    LAEO-Net \cite{marin2019laeo} & -- & -- & -- & 0.80\\
    \hline
    LAEO-Net++ \cite{marin2020laeo} & -- & -- & -- & 0.87\\
    \hline
    \hline
    {\bf Baseline} & 0.77 & 0.80 & 0.78 & 0.86\\
    \hline
    \hline
    {\bf with uncertainty} & 0.80 & 0.72 & 0.76 & {\bf 0.88}\\
    \hline
    \end{tabular}
    \label{tab:socialinteraction}
\end{table}

We report in Tab \ref{tab:socialinteraction} the performance provided by our baseline method and the one incorporating the uncertainty on the test set.
The results suggest that using the prior knowledge derived from the uncertainty allows us to significantly reduce the number of false positive ($-6\%$, with a slight increase of the precision) to the price of a small reduction of true positive ($-7\%$, with a small reduction of the recall). Overall, the uncertainty brings improvements as the AP increases ($+0.02$). As a reference we also show in the table the results provided by \cite{marin2019laeo,marin2020laeo}. 

\begin{figure}[htb]
  \centering
  \includegraphics[width=.8\linewidth]{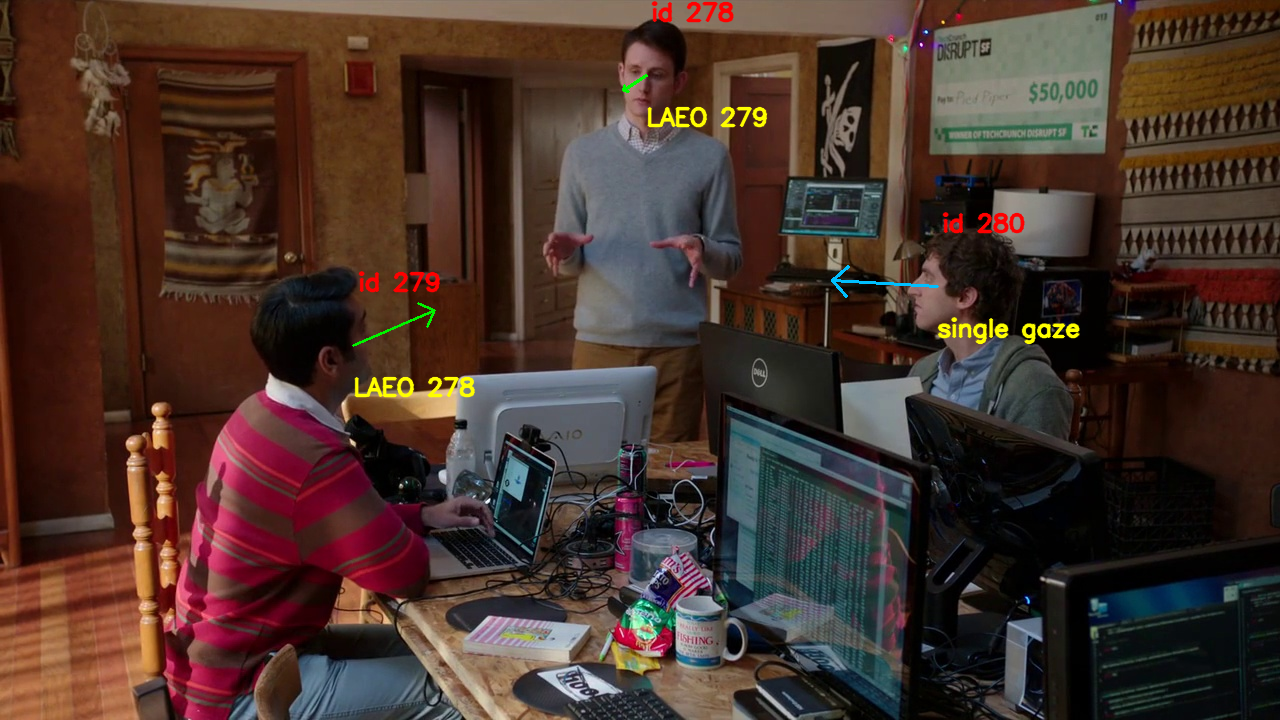}
  \includegraphics[width=.8\linewidth]{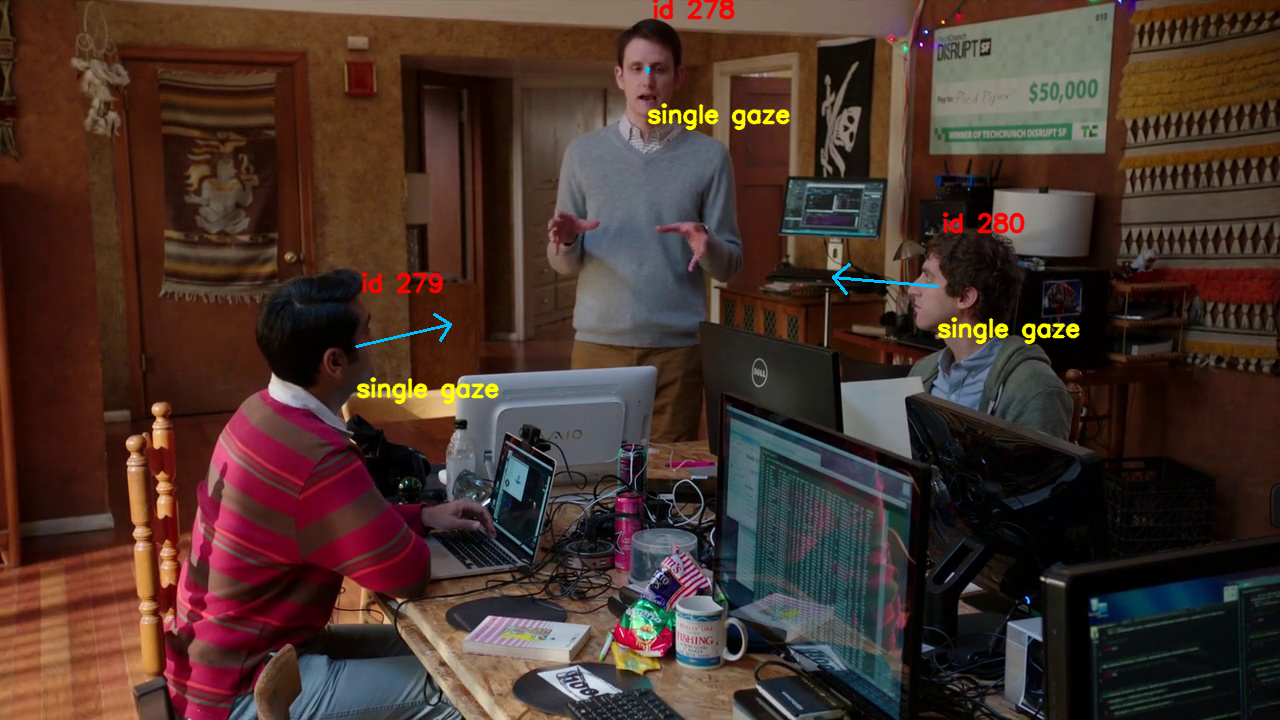} 
  \caption{Examples of LAEO detections. The arrows represent the head direction estimated by HHP-Net and projected on the image plane, and is green if the corresponding person has been found involved in a LAEO. The prediction of our method for LAEO detection is reported in yellow and, in case of LAEO, it specifies the identifier of the other interacting person. The identifiers are in red close the subjects.}
  \label{fig:experimental_figure_1}
\end{figure}

\section{Discussion}
\label{discussion}
In this work we introduced a method for head pose estimation from the head keypoints extracted from RGB images, that provides the head pose as a triplet of Euler angles, each one associated with a measure of the aleatoric heteroscedastic uncertainty. We approached the problem as a multi-task regression, and designed a neural network which is very efficient both in terms of space occupancy (less than 0.5 MB) and of inference time (it runs at 100 fps), thus providing the potential to run on mobile devices. A core element of the architecture is the multi-task loss we employed, in which the data-driven uncertainties act as a weight of the sub-losses, each one related with the estimation of a certain angle and corresponding uncertainty.\\
We provided a thorough experimental assessment, where we discussed the connections between estimation error and uncertainty, that favor the interpretability of our model. We also compared our method with state-of-art approaches, showing comparable or slightly lower performance while providing the lightest method currently available. As an example applications, we discussed the application to social interaction analysis in images.\\
Our work will be extended in different directions. A straightforward one will be to enlarge the number of input points to fully exploit the information provided by the pose detector. This may favor on the one hand the robustness of the method in more challenging situations, and on the other the possibility of considering it as a first step for an activity recognition pipeline, in particular when considering social activities or actions involving the interaction with the environment or with other people. 
With reference to the latter task, we are currently working on the design of more refined measures of social interaction between small groups of people, incorporating more explicitly the uncertainty and the knowledge derived from the 3D head pose estimator, that may help to disambiguate challenging group configurations.  

\section*{Acknowledgement}
	This work has been carried out at the Machine Learning Genoa (MaLGa) center, Università di Genova (IT). It has been supported by Fondazione Cariplo with the grant no. 2018-0858, and by AFOSR,  grant n. FA8655-20-1-7035.

{\small
\bibliographystyle{ieee_fullname}
\bibliography{hhp_net}
}

\clearpage
\newpage\appendix
\begin{center}
{\Large \textbf{Supplementary Material}}

\end{center}

\thispagestyle{empty}


\section{Derivation of our loss function}
\label{derivation_loss}

In this section we report the derivation of the loss function presented in Section \ref{HHP-Net}. 
We target scenarios where uncertainty may be due to data noise and varying on different inputs. Considering a typical regression problem where we want to estimate a function $f_{\omega}$ from the input $x_i$ to the output $y_i$,  we can formalize the setting as
\begin{equation*}
    y_i = f_{\omega} (x_i) + \epsilon(x_i)  
\end{equation*}
where the output can be seen as the sum between a function $f_{\omega}(x_i)$ and $\epsilon(x_i)$ that is the noise that depends on the input $x_{i}$ \cite{nix1994}.

To quantify the uncertainty, the model is trained to learn a function that estimates both the mean and the variance of a target distribution using a maximum-likelihood formulation of a neural network \cite{Buntine1991BayesianB, mackay1992, rumelhart1995}: in order to do that, we need to assume that the errors are normally distributed $\epsilon(x_i) \sim \mathcal{N}(0, \sigma (x_i)^{2})$.

The likelihood for each point $x_i$ is: 

\begin{gather*}
p \left( y_{i} | x_{i}; \omega \right) = \mathcal{N} \left(f_{\omega} \left( x_{i}\right), \sigma \left(x_i \right)^{2} \right) \\
= \frac{1}{\sqrt{2\pi \sigma \left( x_i \right)^{2} }} \exp \left[ -\frac{\left( y_i - f_{\omega}\left(x_i\right)\right)^2}{2\sigma \left( x_i \right)^{2}}\right]
\end{gather*} 
where $y_{i}$ is the mean of this distribution and $\sigma(x_i)^{2}$ is the variance.

The neural network architecture should be modified to include an additional term to the output layer, to predict the variance (or the logarithm of it): this latter quantifies the uncertainty associated with the prediction based on the noise in the training samples (the uncertainty is a function of the input e.g. if the noise is uniform over all the input values, the uncertainty should be constant).

Applying the logarithm to both sides we obtain:

\begin{gather*}
\log p \left( y_{i} | x_{i}; \omega \right) \\
= - \frac{\left( y_i - f_{\omega}\left(x_i\right)\right)^2}{2\sigma \left( x_i \right)^{2}} - \frac{1}{2} \log \sigma \left( x_i \right)^{2}  - \frac{1}{2} \log (2\pi)
\end{gather*} 

that is the log likelihood we want to maximize (the last term is ignored in the following being a constant). 

Maximizing the log likelihood is equivalent to minimizing the negative log likelihood, and therefore we rewrite the minimization problem as:

\begin{equation*}
  \min_{\omega} - \frac{1}{N} \sum_{i=1}^{N} \log p \left( y_{i} | x_{i}; \omega \right)
\end{equation*}

Finally the objective we want to minimize over all $x_i$ becomes:

\begin{equation*}
  \frac{1}{N} \sum_{i=1}^{N} \frac{1}{2\sigma\left( x_{i}\right)^{2}} \left \| y_{i} - f_{\omega} \left( x_{i} \right) \right \|^{2} + \frac{1}{2} \log \sigma \left( x_{i} \right)^{2}
\end{equation*}

In order to solve possible numerical issues the objective is modified in this way:

\begin{equation*}
   \sum_{i=1}^N  \frac{1}{2} \exp \left( -s_{i} \right) \left\| y_{i} - f_{\omega}\left( x_i  \right) \right\|^{2} + \frac{1}{2} s_i  
\end{equation*}

where  $s_{i} = \log \sigma (x_{i})^{2} $: in this way potential divisions by zero are avoided \cite{NIPS2017_2650d608}. 

Lastly we extend this objective for a multi-ouput regression model for training our network obtaining the objective proposed in Section \ref{HHP-Net} of the main paper.

\section{Removing the uncertainty: an ablation study}
\label{remove_unv}

In this section we provide more details about the ablation study discussed in Section \ref{method_assessment} where we  consider two variations of our method:
\begin{description}
\item[MSE:] we  directly regress the three angles adopting a loss computed as the sum of the Mean Squared Error (MSE) on each angle:
\begin{equation*}
 \mathcal{L}_{MSE} =  \sum_{i \in \{y,p,r\}} \left\| q_{i} - f_{i}\left( \mathbf{x}_1, \mathbf{x}_2,\mathbf{c} \right) \right\|^{2}
\end{equation*}
where $\mathbf{q}=[y,p,r]$ ($y$=way,  $p$=pitch and $r$=roll)

\item[COMB:] we employ an alternative loss function $\mathcal{L}_{COMB}$ proposed in \cite{ruiz2018finegrained} which has been proved to be very successful on the same estimation task. The loss allows to jointly solve a regression and classification tasks, and it can be formalized as follows $\mathcal{L}_{COMB}$:

\begin{equation*}
  \sum_{i \in \{y,p,r\}} \sum_{j} -q_{j} \log \left( f_{j} \right) + \alpha * \left\| q_{i} - f_{i}\left( \mathbf{x}_1, \mathbf{x}_2,\mathbf{c} \right) \right\|^{2} 
\end{equation*}

combining the cross entropy loss, computed between the  binned angles, and the MSE loss, computed between the scalar angles; $\alpha$ is an hyperparameter that controls the weight of the regression loss (in the experiment we set $\alpha = 1$).
\end{description}

In Table \ref{tab:lossfunctions} we extend Table \ref{tab:lossfunction} of the main paper and report the angular errors we obtain with the three different losses. As it can be observed, learning the angles associated with the uncertainty provides the best average performance, showing the benefit of the uncertainty not only in terms of interpretability of the model but also as a way to improve its effectiveness.\\

\begin{table}[h]
    \centering
    \caption{Comparison among different losses (see text). {\bf All errors are expressed in degrees ($^\circ$)}:  
    err$_y$= yaw error, err$_p$=pitch error, err$_r$= roll error, MAE = Mean Absolute Error.   }
    \smaller{
    \begin{tabular}{|c|c|c||c|c|c|c|}
    \hline
\textbf{Train} & \textbf{Val} & \textbf{Loss} &  \textbf{err$_y$} & \textbf{err$_p$ } & \textbf{err$_r$ } & \textbf{MAE}\\
    \hline\hline
    BIWI & BIWI & MSE  & 2.90 & 4.80 & 3.34 & 3.70 \\ 
     \hline
    BIWI & BIWI &     COMB & 3.15 & 4.85 & 3.40 & 3.80 \\
     \hline
    BIWI & BIWI &     {\bf  UNC}  & 3.04 & 4.79 & 3.21 & \textbf{3.68} \\ 
     \hline
     \hline
     300WLP & BIWI & MSE & 4.75 &  6.65 &  4.45 & 5.28 \\
     \hline
     300WLP & BIWI & COMB & 4.67 & 8.08 & 4.87 & 5.88 \\ 
     \hline
     300WLP & BIWI & {\bf  UNC} &  4.14 & 7.00 & 4.40 &  {\bf 5.18} \\ 
     \hline
     \hline
     300WLP & AFLW2000 & MSE   & 5.72 & 10.41 & 8.08 & 8.07\\
     \hline
     300WLP & AFLW2000 &COMB & 5.55 & 10.39 & 8.18 & 8.04 \\ 
     \hline
     300WLP & AFLW2000 & {\bf  UNC} &  5.26 & 10.12 & 7.73 & {\bf 7.70} \\ 
     
     \hline
     \hline
     AFLW & AFLW2000 & MSE   & 7.60 & 6.43 &   4.76 &   6.26 \\
     \hline
     AFLW & AFLW2000 &COMB & 7.31 & 6.55 & 4.68 & 6.18 \\ 
     \hline
     AFLW & AFLW2000 & {\bf  UNC} &   7.40 & 6.63 & 4.47 & {\bf  6.16} \\ 
     \hline

    \end{tabular}}
    \label{tab:lossfunctions}
\end{table}

\section{On the number of keypoints}

In this section we provide an extended version of the assessment we discuss in Sec. 4 of the main paper with the aim of observing the influence of the quality and quantity of input semantic features on the final head pose estimate. For the sake of the discussion, we report here plots and comments already included in the main document.

\begin{figure}[h]
    \centering
    \includegraphics[width=.48\linewidth]{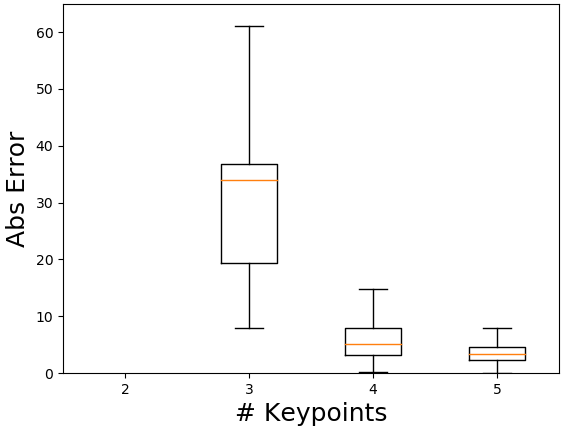}
    \includegraphics[width=.48\linewidth]{images/drop_box_plot_err_no_out_1.png}
    \\
    \includegraphics[width=.48\linewidth]{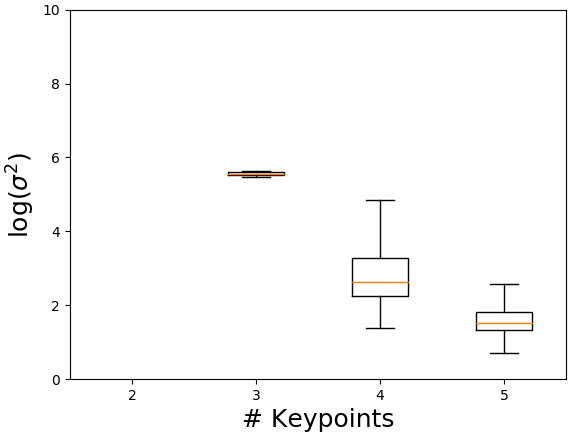}
    \includegraphics[width=.48\linewidth]{images/drop_box_plot_no_out_1.png}
    
    \caption{Performance of our method (top row: mean angular error, bottom row: uncertainty) with respect to the number of input points, considering the output of Open Pose (above) and randomly dropping points from the input (below). We used a model trained on 300W-LP and tested on the whole BIWI (plots refer to the latter).} 
    \label{fig:boxplot_}
\end{figure}

In Figure \ref{fig:boxplot_}, first column, we analyse the performance of our method in terms of uncertainty values (bottom-left) and absolute angular error (top-left) as we group the input data according to the number of keypoints provided by the Open Pose detector. When only 3 keypoints are available the uncertainty is rather high on average. Increasing the number of points it is progressively reduced, with a similar trend shown by the error. This confirms the intuition that the more input points the method has, the higher is its confidence in the prediction, which is more reliable and accurate.\\
Since the worst case scenario corresponds to having at least three keypoints, we randomly dropped points from the input to evaluate the behavior of the method in more challenging scenarios.  The results are shown in the second column of Figure \ref{fig:boxplot_}.
When points are randomly dropped, we only consider samples with more than two points.

When all  the 5 keypoints are available, the uncertainty is compactly lower (confirming what already observed in the previous experiment) as the method can rely on a more comprehensive representation of the input. In the intermediate cases -- where we may have 2, 3, or 4 keypoints available in input -- the uncertainty progressively decreases, but we also have a higher degree of variability, as some keypoints configurations are more significant than others and thus the amount of information they provide to the model may be uneven reflecting the concept that the noise could be different for each input sample. With respect to the plots in the first column of Figure \ref{fig:boxplot}, the box plots at right show a higher standard deviation since randomly dropping points from the input we simulate a higher variability in the input configurations with respect to the ones usually provided by Open Pose and from the datasets we used.\\

\section{Model size and parameters}

In this section we show the robustness of our method with respect to reductions of size, that may be needed when the available computational resources are very limited.
More specifically, we analyse how the performance changes as we reduce the size of the model.  
We choose 300W-LP training and BIWI test (protocol P1) for its larger training and test sets and decrease the number of neurons in the fully connected layers so the backbone remains the same proposed in the paper, while its size decreases. Given a reduction factor $\alpha \in (0, 1)$, we obtain a ``reduced" version of  our architecture by multiplying the original number of neurons in each layer (250, 200 and 150 in, respectively, the first, second and third layer) by $\alpha$. 

By varying  $\alpha$ in the range $(0,1)$ we reduce the model size (the number of parameters) and thus also the number of sum and multiplication operations. Table \ref{tab:4} compares our baseline ($\alpha=1$) with two reduced models (overall size in MB up to $10 \times$ smaller) causing a very limited degradation in the Mean Absolute Error (below 1 degree). This experiment highlights the possibility of further reducing the size of the architecture, with a very limited performance loss, if required by the system.

\begin{table}[H]
  \caption{Comparison among models with different sizes (Protocol P1: 300W-LP train,  BIWI test). $\alpha$ = neurons reduction factor (see text), MAE = Mean Absolute Error}
  \label{tab:4}
  \centering
  \begin{tabular}{|c || c | c | c | c |}
    \hline
    \textbf{$\alpha$} & \textbf{MAE} & \textbf{Mult-adds} & \textbf{Parameters} & \textbf{MB} \\

    \hline\hline
    
     1 & 5.18 & 93000 & 94000 & 0.385 \\ 
     \hline
     0.6 & 5.43 & 37000 & 37000 & 0.158 \\ 
     \hline
     0.2 & 5.54 & 6000 & 6000 & 0.032 \\ 
     \hline

     \hline
\end{tabular}
\end{table}

\end{document}